\documentclass{article}
\usepackage{ijcai26}

\usepackage{times}
\usepackage{soul}
\usepackage{url}
\usepackage[hidelinks]{hyperref}
\usepackage[utf8]{inputenc}
\usepackage[small]{caption}
\usepackage{graphicx}
\usepackage{amsmath}
\usepackage{amsthm}
\usepackage{booktabs}
\usepackage{algorithm}
\usepackage{algorithmic}
\usepackage[table]{xcolor}
\usepackage[switch]{lineno}
\usepackage{listings}
\usepackage{scalerel}
\usepackage{placeins}
\definecolor{codegreen}{rgb}{0,0.6,0}
\definecolor{codegray}{rgb}{0.5,0.5,0.5}
\definecolor{codepurple}{rgb}{0.58,0,0.82}
\definecolor{backcolour}{rgb}{0.98,0.98,0.98}

\title{LELA: An End-to-end LLM-based Entity Linking Framework\\
with Zero-shot Domain Adaptation}

\author{
    Samy Haffoudhi \and Nikola Dobričić \and Fabian Suchanek \and Nils Holzenberger
    \affiliations
    Télécom Paris, Institut Polytechnique de Paris
    \emails
    \{samy.haffoudhi, nikola.dobricic, fabian.suchanek, nils.holzenberger\}@telecom-paris.fr
}

\begin{document}

\maketitle

\begin{abstract}
    Entity linking is a key component of many downstream NLP systems, yet existing approaches are often tied to the specific target knowledge bases and domains, limiting their real world application. In this paper, we extend LELA, a modular and domain-agnostic LLM-based entity disambiguation method, into a practical Python library that integrates zero-shot Named Entity Recognition (NER) -- thereby providing a complete end-to-end pipeline for entity-linking in real-world usage. We provide experimental results validating LELA's performance and robustness across diverse entity linking settings. 
In our demo, users can play with the system on their own input texts. All code is publicly available at \url{https://github.com/NDobricic/LELA}, and a video is at \url{https://www.youtube.com/watch?v=WdupiRjLbR4}.
\end{abstract}

\section{Introduction}

Entity linking (EL) is the task of identifying and mapping ambiguous mentions of entities in a natural language text to reference entities in a knowledge base (KB). 
For example, the text can be:\\[-2mm]

\texttt{France hosted the Olympics in Paris.\\[-2mm]} 

\noindent The KB contains each entity, associated with a short textual description, e.g.:\\[-2mm]

    \ttfamily
\begin{tabular}{ll}
Paris (city)& Capital city of France\\
Paris (novel)& 1897 novel by Emile Zola\\
France & Country in Europe\\
France Gall & French singer\\
\end{tabular}
\rmfamily
\ \\[0mm]
\noindent Entity linking consists of two sub-tasks, \emph{mention detection} (MD) and \emph{entity disambiguation} (ED). Mention detection aims to identify the mentions of entities (here: ``France'', ``the Olympics'', and ``Paris''). Mention detection often boils down to Named Entity Recognition (NER). Entity disambiguation, on the other hand, aims to map the identified mentions to their correct entity in the KB, if they exist (here: ``France'' to \texttt{France} and ``Paris'' to \texttt{Paris (city)}, with no mapping for ``Olympics''). Entity Linking is an important preprocessing step in tasks such as information extraction~\cite{martinez-rodriguezInformationExtractionMeets2020}, knowledge-based question answering~\cite{weltyComparisonHardFilters2012}, and knowledge graph completion~\cite{jiKnowledgeBasePopulation}. 

\begin{figure}
\centering
\begin{lstlisting}[escapechar=?]
from lela import Lela

# Choose each component of LELA
config = {
  "loader": {
    "name": "text" # or: pdf, docx, ...
  },
  "ner": {
    "name": "gliner", # or: regex, spacy
    "params": {"labels": ["person", "location"]},
  },
  "candidate_generator": {"name": "bm25"}, 
  ?{\hspace{2em}}?# or: fuzzy, dense, openai_api_dense
  "reranker": {"name": "llama_server"}, 
  ?{\hspace{2em}}?# or: none, cross_encoder_vllm...
  "disambiguator": {
    "name": "vllm", # or: first, openai_api, transformers
    "params": {"model_name": "Qwen/Qwen3-4B"},
  },
  "knowledge_base": {
    "name": "jsonl", "params": {"path": "my_kb.jsonl"}
  },
}

lela = Lela(config)

# Run the pipeline on a document
results = lela.run("docs/file1.txt")
\end{lstlisting}
\vspace{-5mm}
\caption{LELA is designed for a modular use in Python.}
\label{fig:el_pipeline}
\vspace{-\baselineskip}
\end{figure}

Most entity linking approaches focus on linking to general KBs such as Wikidata~\cite{vrandecicWikidataNewPlatform2012}, DBpedia~\cite{auerDBpediaNucleusWeb2007} or Yago~\cite{suchanek_yago_2024}. However, in real-world applications, the KB is often proprietary or domain-specific, as in the legal or biomedical domain, or inside a company. Entity disambiguation for such KBs has been addressed recently by the LELA system~\cite{haffoudhi_lela_2026}, which can disambiguate entities in a true zero-shot fashion without any need for training data or fine-tuning.
However, LELA performs only entity disambiguation, not mention detection, i.e., it assumes that entity mentions have already been identified in the text. In real-world scenarios, however, the mentions of the entities are not marked and have to be detected.

In this demo paper, we extend LELA to an end-to-end EL system that can ingest text and a KB and output a disambiguation -- with no need for either training-data, fine-tuning, or the prior identification of the mentions. We contribute a modular and fully open framework, where different named-entity extractors, candidate retrievers, rerankers, and disambiguators can be used interchangeably. 
Our demo lets users run LELA on their own input texts with different KBs.

\section{Related work}

Current research typically bifurcates the problem into two distinct tasks: zero-shot NER, which identifies mentions of unseen entity types~\cite{zaratiana_gliner_2024,bogdanov_nuner_2024,cocchieriZeroNERFuelingZeroShot2025}, and zero-shot ED, which maps mentions to a KB not encountered during training~\cite{logeswaranZeroShotEntityLinking2019,wuScalableZeroshotEntity2020,haffoudhi_lela_2026}. To our knowledge, a unified, end-to-end architecture for zero-shot EL remains unexplored. Consequently, existing frameworks such as spaCy\footnote{https://spacy.io/api/large-language-models\#el-v1}, Zshot~\cite{picco_zshot_2023}, and GLINKER\footnote{a recently-proposed production-oriented framework built around GLiNER~\cite{stepanovGLiNERMultitaskGeneralist2024}: \url{https://github.com/Knowledgator/GLinker?tab=readme-ov-file}} rely on modular, multi-stage pipelines. 
However, these are often tightly coupled to specific architectures.
LELA differentiates itself by offering a flexible LLM-native architecture and integrating the state-of-the-art MD and ED components, leading to unparalleled robustness in real-world scenarios. Our experiments confirm that the capabilities of LLMs are essential for handling the ambiguity of true zero-shot ED, allowing LELA to significantly outperform existing frameworks on complex, unseen domains.

\section{System design}

Our work extends LELA from a disambiguation-only method to a complete end-to-end entity linking framework. Our system is built as a modular pipeline on top of spaCy's component architecture, where each stage is a pluggable component that can be changed independently. The pipeline works in four stages:

\paragraph{1. Named Entity Recognition.}
The first stage identifies entity mentions in the input text. We provide several interchangeable NER components: a zero-shot GLiNER model~\cite{zaratiana_gliner_2024} (default: NuNER\_Zero-span~\cite{bogdanov_nuner_2024}), spaCy's pretrained NER models,
and a lightweight regex-based recognizer for rapid prototyping or well-defined mention formats. Long documents are automatically chunked with overlap to respect model context limits. 

\paragraph{2. Candidate Generation.}
For each detected mention, the system retrieves candidate entities from the knowledge base. Available retrievers include: BM25 (with rank-bm25\footnote{\url{https://github.com/dorianbrown/rank_bm25/tree/master}}), dense retrieval with FAISS\footnote{\url{https://github.com/facebookresearch/faiss}} indexing, and fuzzy string matching (via RapidFuzz\footnote{\url{https://github.com/rapidfuzz/RapidFuzz}}).

\paragraph{3. Reranking (optional).}
Candidates can be reranked to filter the set passed to the disambiguator using cross-encoder models with top-$k$ cutoff. This stage can also be skipped.
See \cite{haffoudhi_lela_2026} for a full study of the use of reranking for entity disambiguation.

\paragraph{4. Disambiguation.}
The final stage selects the correct entity from the candidates using LLM reasoning, following the LELA methodology~\cite{haffoudhi_lela_2026}. The LLM receives the top-$k$ candidates from the reranker along with the input context (with the mention marked in square brackets) and selects the most likely entity by index.
LLM-based disambiguation supports linking a detected mention to ``NIL'' via an explicit ``None of the candidates'' option. A baseline disambiguator that selects by retrieval rank is also provided. The entity is then displayed when hovering over the highlighted span.

\newcommand{\hflogo}{\scalerel*{\includegraphics[trim=40 40 40 40]{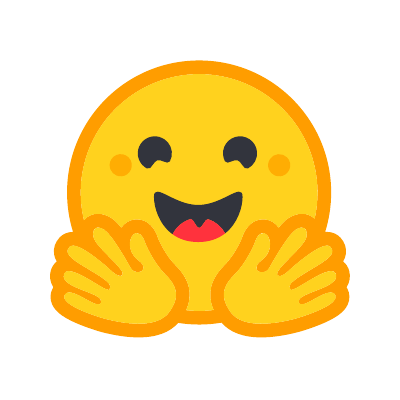}}{X}}

\paragraph{Backends.} The components 2-4 can also be disabled for use cases that require only mention detection. For a given component, the framework provides 
different backend options: vLLM\footnote{\url{https://github.com/vllm-project/vllm}} for fast LLM inference using the offline API, the \hflogo~Transformers\footnote{\url{https://github.com/huggingface/transformers}} library, SentenceTransformers\footnote{\url{https://github.com/huggingface/sentence-transformers}} for the embedding and reranking models, as well as HTTP requests to OpenAI API compatible endpoints\footnote{\url{https://developers.openai.com/api/reference/overview}}, for interacting with models served using vLLM (online API), llama.cpp\footnote{\url{https://github.com/ggml-org/llama.cpp}} (for quantized models, potentially running on CPUs), or proprietary models hosted remotely. 

\paragraph{Interfaces.} The KB is given as a simple JSONL file, with each line representing an entity with an identifier, a label, and a description.
Our system can be used via three interfaces:
\begin{itemize}
    \item a \textbf{Python API} for programmatic use (as in Figure~\ref{fig:el_pipeline}),
    \item a \textbf{command-line interface (CLI)} (\texttt{lela -{}-config config.json -{}-input doc.txt}) for batch processing of documents in various formats (text, PDF, DOCX, HTML, JSON, JSONL), and
    \item an interactive \textbf{Web application} built with Gradio\footnote{\url{https://www.gradio.app/}} (described in Section~\ref{sec:demo}).
\end{itemize} 
All three interfaces share the same underlying spaCy pipeline and configuration system. The system supports caching, progress tracking, and GPU memory estimation to help users select appropriate model configurations.

\paragraph{Extensibility.}
New components can be added by registering a spaCy factory (for pipeline stages) or a registry entry (for loaders and knowledge bases). Each component follows a simple protocol:
(1)~NER components populate \texttt{doc.ents}, (2)~candidate generators populate a custom \texttt{candidates} extension, and (3)~disambiguators set a \texttt{resolved\_entity} extension -- making it straightforward to integrate domain-specific models without modifying the core framework.

\section{Experimental Validation}

\paragraph{Datasets.} We evaluate our framework on two different benchmark datasets:
 Elgold~\cite{islamaj_nlm-gene_2021}, which evaluates entity linking to Wikipedia across seven domains; and MHERCL~\cite{graciotti_musical_2025} which focuses on Wikidata long-tail entities and the musical heritage domain.

\paragraph{LELA configuration.} We report results for LELA using the \texttt{Qwen3-Embedding-4B} retriever, the \texttt{Qwen3-Reranker-4B} reranker~\cite{zhang_qwen3_2025} and the \texttt{Qwen3-30B-A3B} reasoning LLM~\cite{yang_qwen3_2025}. We retrieve 100 candidates per mention, set the top-$k$ cutoff to 10 candidates, and sample three outputs for self-consistency. We use the \texttt{NuNER\_Zero-span} model~\cite{bogdanov_nuner_2024} for mention-detection with a fixed NER label set, based on the set of mention types present in the benchmark.

\paragraph{Baselines.} We evaluate BLINK~\cite{wu_scalable_2020}, as another representative of zero-shot entity disambiguation methods, and the two state-of-the-art end-to-end entity linking methods for Wikipedia: ReFinED~\cite{ayoola_refined_2022} and Relik~\cite{orlando_relik_2024}. On MHERCL, we also compare to the recently-proposed GLiNKER framework.

\paragraph{Evaluation Metrics.} We measure performance using ELEVANT~\cite{bast_elevant_2022}. On Elgold, we report \textit{InKB} EL F1 score~\cite{roder_gerbil_2018}. On MHERCL, we follow prior work~\cite{graciotti_musical_2025} and report EL F1.

\paragraph{Results.} On the \textbf{Elgold} benchmark (Table~\ref{tab:elgold_el_only}), LELA performs on par with the fully supervised state-of-the-art models, ReFinED and Relik. LELA achieves this competitiveness without the required extensive task-specific training on Wikipedia. This advantage is most visible in the complex domains such as the \textit{Science paper abstracts}, where LELA surpasses ReFinED by nearly 18 percentage points, demonstrating superior robustness.
Furthermore, LELA systematically outperforms BLINK, when both rely on the same MD component, confirming the performance gains induced by the reasoning capabilities of our LLM-based ED compared to standard encoder-only baselines.

On the \textbf{MHERCL} benchmark (Table~\ref{tab:mhercl}), which targets long-tail entities in the musical domain, LELA establishes a new state-of-the-art. It outperforms ReFinED and significantly surpasses other zero-shot or modular baselines. These results confirm that while supervised methods struggle to adapt to specialized domains without retraining, LELA handles arbitrary distributions robustly.

\newcommand{\conf}[1]{\textcolor{black}{\fontsize{6}{6}\selectfont$\pm$#1}}
\newcommand{\res}[2]{#1\conf{#2}}
\newcommand{\bres}[2]{\textbf{#1}\conf{#2}}
\newcommand{\istop}[1]{\cellcolor{lightgray} #1}

\begin{table}
  \centering
  \small 
  \setlength{\tabcolsep}{3pt}
  \begin{tabular}{lcccc}
    \toprule
    \textbf{Domain} & \textbf{BLINK+NER} & \textbf{Relik} & \textbf{ReFinED} & \textbf{LELA} \\
    \midrule
    \textbf{1 (News)} & \res{68.2}{3.1} & \istop{\res{76.5}{2.9}} & \istop{\bres{78.4}{2.8}} & \res{74.7}{3.0} \\
    \textbf{2 (Jobs)} & \res{43.3}{4.2} & \res{67.0}{5.1} & \istop{\bres{72.0}{4.6}} & \res{60.2}{4.7} \\
    \textbf{3 (Movie)} & \res{69.2}{4.8} & \istop{\res{72.5}{5.0}} & \istop{\res{74.3}{4.8}} & \istop{\bres{75.3}{4.7}} \\
    \textbf{4 (Auto)} & \res{63.7}{5.4} & \res{65.8}{6.0} & \istop{\bres{74.9}{5.1}} & \res{66.5}{5.6} \\
    \textbf{5 (Amazon)} & \res{63.4}{4.7} & \istop{\res{67.8}{4.9}} & \istop{\res{66.9}{4.9}} & \istop{\bres{71.5}{4.6}} \\
    \textbf{6 (Science)} & \res{29.3}{3.3} & \res{31.5}{4.0} & \res{23.8}{3.8} & \istop{\bres{41.6}{3.8}} \\
    \textbf{8 (Historic)} & \istop{\res{65.7}{7.0}} & \istop{\res{68.2}{7.4}} & \istop{\bres{72.2}{6.9}} & \istop{\res{69.9}{7.0}} \\
    \midrule
    \textbf{Macro} & \res{57.6}{1.8} & \res{64.2}{2.0} & \istop{\bres{66.1}{1.8}} & \istop{\res{65.7}{1.8}} \\
    \bottomrule
  \end{tabular}
  \caption{F1 values (InKB, in \%) on Elgold across the seven domains, with 95\% confidence intervals. Gray results have overlapping confidence intervals with the best results.}
  \label{tab:elgold_el_only}
\end{table}

\begin{table}
  \centering
  \small
  
  \begin{tabular}{lc}
    \toprule
    \textbf{Method} & \textbf{EL F1} \\
    \midrule

    GLiNKER~\cite{stepanovGLiNERMultitaskGeneralist2024} & 16\conf{1.5} \\
    BLINK+NER~\cite{wuScalableZeroshotEntity2020} & 47\conf{2.0} \\
    Relik~\cite{orlando_relik_2024} & 44\conf{2.0} \\
    ReFinED~\cite{ayoola_refined_2022} & 49\conf{2.0} \\
    \textbf{LELA (ours)} & \textbf{56}\conf{2.0} \\
    \bottomrule
  \end{tabular}
  \caption{F1 values (in \%) on the MHERCL benchmark, with 95\% confidence intervals.}
  \label{tab:mhercl}
\end{table}

\section{Demo}
\label{sec:demo}

Our demo allows users to link entities in arbitrary documents with different KBs using LELA's Web application. Our interface shows the different steps of the disambiguation process and measures the execution time. Finally, it shows the recognized mentions with their link to the entities in the KB, or to NIL. The Web interface mirrors the full modularity of our approach: Users can experiment with different NER systems, candidate retrievers, rerankers, and reasoning LLMs -- trading off speed against performance. We provide a large range of reference KBs to try out: Users can work with both general-purpose KBs (such as YAGO~\cite{suchanek_yago_2024}) and domain-specific KBs (such as the Crossref Funder Registry\footnote{https://www.crossref.org/services/funder-registry/} or the 16 different Wikia domains from ZESHEL~\cite{logeswaran_zero-shot_2019}). 

\textbf{In a twist that goes beyond existing demos, users can come up with their own KBs}. For example, a user interested in biology can create a small dictionary with two meanings of the word ``culture'' (the process of growing cells in the lab vs. the ensemble of arts, customs, and traditions), and ask the system to identify mentions to these entities and determine which meaning is intended in a sentence. Similarly, users from industry can create a dictionary of technical terms that their own approaches struggle with (such as terms that have specific meanings in industry jargon), and see if our approach copes, experimenting with different pipeline configurations.

\section{Conclusion}
\label{sec:conclusion}

LELA is a Python library for entity linking, handling both mention detection and entity disambiguation. It is true zero-shot, in the sense that it needs neither training data nor fine-tuning, and works on domain-specific text and knowledge bases out-of-the-box. LELA can be used through a Python API, a CLI or a Web Interface, making it suited for real-world applications. It is modular, offering different options for the various stages of the entity linking pipeline, and can be extended with additional components. 
Future work can integrate additional components, add support for KBs in Turtle format, as well as automate the generation of textual descriptions and NER labels.

\section*{Acknowledgements}

The work was partially supported by Agence de l'Innovation de Défense – AID - via Centre Interdisciplinaire d’Etudes pour la Défense et la Sécurité – CIEDS - (project 2024 - KB-LM).

\FloatBarrier

\bibliographystyle{named}
\bibliography{my_library}

\end{document}